\pdfoutput=1
\documentclass[11pt,a4paper]{article}
\usepackage[hyperref]{emnlp2020}
\usepackage{times}
\usepackage{latexsym}

\usepackage{microtype}

\aclfinalcopy 

\usepackage{soul}      
\usepackage{colortbl}  
\usepackage{amsmath}   
\usepackage{amsfonts}  
\usepackage{amssymb}   
\usepackage{enumitem}
\usepackage{graphicx}
\usepackage{multirow}
\usepackage{colortbl}
\usepackage{rotating, graphicx}
\usepackage{multirow, makecell, caption}
\usepackage{booktabs}
\usepackage{url}

\usepackage{amsmath}
\makeatletter
\newcommand*\@bigplus[1]{\vcenter{\hbox{#1$\m@th +$}}}
\newcommand*\bigplus{%
    \DOTSB 
    \mathop{%
        \mathchoice
            {\@bigplus \huge}%
            {\@bigplus \LARGE}%
            {\@bigplus {}}%
            {\@bigplus \footnotesize}%
    }%
    \slimits@ 
}
\makeatother

\title{AVA: an Automatic eValuation Approach to Question Answering Systems}

\author{Thuy Vu \\
  Amazon Alexa\\
  Manhattan Beach, CA, USA \\
  \texttt{thuyvu@amazon.com} \\\And
  Alessandro Moschitti \\
  Amazon Alexa\\
  Manhattan Beach, CA, USA \\
  \texttt{amosch@amazon.com} \\}

\date{}

\definecolor{mygreen}{rgb}{0.0, 0.44, 0.0}
\newcommand{\AVA}{AVA}
\newcommand{\qaNQ}{AS2-NQ}
\newcommand{\qaGPD}{AS2-GPD}
\newcommand{\avaNQ}{AVA-NQ}

\newcommand{\avaGPD}{AVA-GPD}
\newcommand{\avaADS}{ADS}
\newcommand{\ASS}{AS2}

\begin{document}
\maketitle


\begin{abstract}


We introduce {\AVA}, an automatic evaluation approach for Question Answering, which given a set of questions associated with Gold Standard answers, can estimate system Accuracy.
%
{\AVA} uses Transformer-based language models to encode question, answer, and reference text.
This allows for effectively measuring the similarity between the reference and an automatic answer, biased towards the question semantics.
%
%
%
To design, train and test {\AVA}, we built multiple large training, development, and test sets on both public and industrial benchmarks. Our innovative solutions achieve up to 74.7\% in F1 score in predicting human judgement for single answers. 
Additionally, {\AVA} can be used to evaluate the overall system Accuracy with an RMSE, ranging from 0.02 to 0.09, depending on the availability of multiple references.
\end{abstract}

\section{Introduction}

Accuracy evaluation is essential both to guide system development as well as to estimate its quality, which is important for researchers, developers, and users. This is often conducted using benchmarking datasets, containing a data sample, \emph{possibly} representative of the target data distribution, provided with Gold Standard (GS) labels (typically produced with a human annotation process). The evaluation is done by comparing the system output with the expected labels using some metrics.

This approach unfortunately falls short when dealing with generation tasks, for which the system output may span a large, possibly infinite, set of correct items. For example, in case of Question Answering (QA) systems, the correct answers for the question, \emph{Where is Rome located ?} is large. As it is impossible, also for cost reasons, to annotate all possible system pieces of output, the standard approach is to manually re-evaluate the new output of the system. This dramatically limits the experimentation velocity, while increasing significantly the development costs. 

Another viable solution in specific domains consists in automatically generating an evaluation score between the system and the reference answers, which correlates with human judgement. The BLEU score, for example, is one popular measure in Machine Translation~\cite{papineni-etal-2002-bleu}. This, however, can only be applied to specific tasks and even in those cases, it typically shows limitations~\cite{DBLP:journals/corr/abs-1803-08409}.
As a consequence there is an active research in learning methods to automatically evaluate MT systems~\cite{ma-etal-2019-results}, while human evaluation becomes a requirement in machine translation benchmarking~\cite{barrault-etal-2019-findings}.

QA will definitely benefit by a similar approach but the automatic evaluation is technically more complex for several reasons:
First, segment overlapping metrics such as BLEU, METEOR, or ROUGE, do not work since the correctness of an answer loosely depends on the match between the reference and candidate answers. For example, two text candidates can be correct and incorrect even if they only differ by one word (or even one character), e.g., for the questions, \emph{Who was the 43$^{rd}$ president of USA ?}, a correct answer is \emph{George W. Bush}, while the very similar answer, \emph{George H. W. Bush}, is wrong.

Second, the matching between the answer candidates and the reference must be carried out at semantic level and it is radically affected by the question semantics. For example, \emph{match}$(t, r | q_1)$ can be true but \emph{match}$(t, r | q_2)$ can be false, where $t$ and $r$ are a pair of answer candidate and reference, and $q_1$ and $q_2$ are two different questions. 
This can especially happen for the case of the so-called non-factoid questions, e.g., asking for a description, opinion, manner, etc., which are typically answered by a fairly long explanatory text. For example, Table~\ref{nonfactq} shows an example of a non factoid question and three different valid answers, which share similarity with respect to the question. However, if the question were, \emph{what may cause anxiety ?}, {Answer 1} and {Answer 3} would intuitively look less related to {Answer 2}.

\begin{table}
\small
\centering
\resizebox{\linewidth}{!}{%
\begin{tabular}{|p{7.5cm}|}
\hline
\textbf{Question}: What does cause left arm pain ?\\ 
\textbf{Reference}: Arm pain can be caused by a wide variety of problems, ranging from joint injuries to compressed nerves; if it radiates into your left arm can even be a sign of a heart attack.\\ 
\hline
\hline
\textbf{Answer 1}: It is possible for left arm pain to be caused from straining the muscles of the arm, pending heart attack, or it can also be caused from indigestion.\\ 
\hline
\textbf{Answer 2}: Anxiety can cause muscles in the arm to become tense, and that tension could lead to pain.\\
\hline
\textbf{Answer 3}: In many cases, arm pain actually originates from a muscular problem in your neck or upper spine. \\
\hline
\end{tabular}
}
\caption{Example of a non-factoid questions}
\label{nonfactq}
\end{table}

In this paper, we study the design of models for measuring the Accuracy of QA systems. In particular, we design several pre-trained Transformer models~\cite{devlin2018bert,DBLP:journals/corr/abs-1907-11692} that encode the triple of question $q$, candidate $t$, and reference $r$ in different ways.

Most importantly, we built (i) two datasets for training and testing the point-wise estimation of QA system output, i.e., the evaluation if an answer is correct or not, given a GS answer; and (ii) two datasets constituted by a set of outputs from several QA systems, for which {\AVA} is supposed to estimate the Accuracy.

The results show a high Accuracy for point-wise models, up to 75\%. Regarding the overall Accuracy estimation, {\AVA} can almost always replicate the ranking of systems in terms of Accuracy performed by humans. Finally, the RMSE with respect to human evaluation depends on the datasets, ranging from 2\% to 10\%, with an acceptable Std. Dev. lower than 3-4\%.

The structure of the paper is as follows: we begin with the description of the problem in Sec.~\ref{sec:problem}.
This is then followed by the details of the data construction and model design, which are key aspects for  system development, in sections~\ref{sec:model} and~\ref{sec:data}.
We study the performance of our models in three different evaluation scenarios in Sec.~\ref{sec:experiments}.

\section{Related Work}

Automatic evaluation has been an interesting research for decades~\cite{papineni-etal-2002-bleu,Magnini2002TowardsAE}.
There are two typical strategies to design an automatic evaluator: supervised and unsupervised.
In machine translation, for example, BLEU~\cite{papineni-etal-2002-bleu} has been a very popular unsupervised evaluation method for the task.
There are also other supervised methods recently proposed, most notably~\cite{ma-etal-2019-results}.
For dialog systems, neural-based automatic evaluators are also studied~\cite{DBLP:journals/corr/abs-1904-10635,lowe-etal-2017-towards,DBLP:journals/corr/TaoMZY17,DBLP:journals/corr/KannanV17}

QA has been traditionally studied early in literature~\cite{Green:1961:BAQ:1460690.1460714}.
QA has recently been used to evaluate a summarization task~\cite{eyal-etal-2019-question}.
Automatic evaluation for QA was addressed by~\citet{Magnini2002TowardsAE} and also for multiple subdomain QA systems~\cite{Leidner-CallisonBurch:2003:CLUK,Lin2006MethodsFA,Shah:2010:EPA:1835449.1835518,Gunawardena2015PerformanceET}.
However, little progress has been made in the past two decades towards obtaining a standard method.
Automating QA evaluation is still an open problem and there is no recent work supporting it.



\section{Problem Definition}
\label{sec:problem}
We target the automatic evaluation of QA systems, for which system Accuracy (the percentage of correct answers) is the most important measure. We also consider more complex measures such as MAP and MRR in the context of  Answer Sentence Reranking/Selection.
\subsection{Answer Sentence Selection (AS2)}
The task of reranking answer sentence candidates provided by a retrieval engine can be modeled with a classifier scoring the candidates.
Let $q$ be a question, $T_q=\{t_1, \dots, t_n\}$ be a set of answer sentence candidates for $q$, we define $\mathcal{R}$ as a ranking function, which orders the candidates in $T_q$ according to a score, $p\left(q, t_i\right)$, indicating the probability of $t_i$ to be a correct answer for $q$.
Popular methods modeling $\mathcal{R}$ include Compare-Aggregate~\cite{DBLP:journals/corr/abs-1905-12897}, inter-weighted alignment networks~\cite{shen-etal-2017-inter}, and BERT~\cite{garg2019tanda}.

\begin{table}
\small
\centering
\resizebox{\linewidth}{!}{%
\begin{tabular}{|lp{7.2cm}|}
\hline
$q$: & What is the population of California?\\
$r$: & With slightly more than 39 million people (according to 2016 estimates), California is the nation's most populous state—its population is almost one and a half times that of second-place Texas (28 million).\\
$s$: & 39 million\\
$t$: &The resident population of California has been steadily increasing over the past few decades and has increased to 39.56 million people in 2018.\\
\hline
\end{tabular}
}
\caption{An example of input data}
\label{evaluator-input}
\end{table}

\subsection{Automatic Evaluation of QA Accuracy}
The evaluation of system Accuracy can be approached in two ways:
(i) evaluation of the single answer provided by the target system, which we call point-wise evaluation; and (ii) the aggregated evaluation of a set of questions, which we call system-wise evaluation.
 
We define the former as a function: $\mathcal{A}\left(q, r, t_i\right) \rightarrow \{0, 1\}$, where $r$ is a reference answer (GS answer) and the output is simply a correct/incorrect label.
%
Table~\ref{evaluator-input} shows an example question associated with a reference, a system answer, and a short answer\footnote{The latter can be very effective but it adds an additional annotation cost, thus we limit its use just for the baseline model. That is, we aim to have a lower cost {\AVA} model}.

A configuration of $\mathcal{A}$ is applied to compute the final Accuracy of a system using an aggregator function. In other words, to estimate the overall system Accuracy, we simply assume the point-wise AVA predictions as they were the GS.
For example, in case of the Accuracy measure, we simply average the AVA predictions, i.e., $\frac{1}{|Q|}\sum_{q\in Q} \mathcal{A}(q,r,t_i[,s])$, where $s$ is a short answer (e.g., used in machine reading). It is an optional input, which we only use for a baseline, described in Section~\ref{sec:linear}.

\section{Model for AVA}
\label{sec:model}

The main intuition on building an automatic evaluator for QA is that the model should capture (i) the same information a standard QA system uses; while (ii) exploiting the semantic similarity between the system answer and the reference, biased by the information asked by the question. 
We build two types of models: (i) linear classifier, which is more interpretable and can help us to verify our design hypothesis and (ii) Transformer-based methods, which have been successfully used in several language understanding tasks.

\subsection{Linear Classifier}
\label{sec:linear}
Given an input example, $\left(q, r, s, t\right)$, our classifier uses the following similarity features:
$x_1$=\emph{sim-token}$\left(s,r\right)$, $x_2$=\emph{sim-text}$\left(r,t\right)$, $x_3$=\emph{sim-text}$\left(r,q\right)$; and $x_4$=\emph{sim-text}$\left(q,t\right)$, where \emph{sim-token} between $s$ and $r$ is a binary feature testing if $r$ is included in $s$, $\emph{sim-text}$ is a sort of Jaccard similarity:
$$\emph{sim-text}\left(s_i, s_j\right)= 2\frac{|\text{\emph{tok}}\left(s_i\right)\cap\text{\emph{tok}}\left(s_j\right)|}{|\text{\emph{tok}}\left(s_i\right)| + |\text{\emph{tok}}\left(s_j\right)|},$$
and $\emph{tok}\left(s\right)$ is a function that splits $s$ into tokens.

Let ${\bf x} = f\left(q,r,s,t\right)=\left(x_1, x_2, x_3, x_4\right)$ be a similarity feature vector describing our evaluation tuple. We train ${\bf w}$ on a dataset $D=\{d_i:\left({\bf x}_i,l_i\right)\}$ using SVM, where $l_i$ is a binary label indicating whether $t$ answers $q$ or not.
We compute the point-wise evaluation of $t$ as the test ${\bf x}\!\cdot\!{\bf w} > \alpha$, where $\alpha$ is a threshold trading off Precision for Recall in standard classification approaches.

\subsection{Transformer-based models}

Transformer-based architectures have proved to be powerful language models, which can capture complex similarity patterns. Thus, they are suitable methods to improve our basic approach described in the previous section. Following the linear classifier modeling,
we propose three different ways to exploit the relations among the members of the tuple $\left(q,r,s,t\right)$.

Let $\mathcal{B}$ be a pre-trained language model, e.g., the recently proposed BERT~\cite{devlin2018bert}, RoBERTa~\cite{DBLP:journals/corr/abs-1907-11692}, XLNet~\cite{DBLP:journals/corr/abs-1906-08237}, AlBERT~\cite{anonymous2020albert}.
We use a language model to compute the embedding representation of the tuple members: $\mathcal{B}\left(a, a'\right) \rightarrow {\bf x} \in \mathbb{R}^d$,
where $\left(a, a'\right)$ is a sentence pair, ${\bf x}$ is the output representation of the  pair, and $d$ is the dimension of the output representations.
The classification layer is a standard feedforward network as 
$\mathcal{A}\left({\bf x}\right) = {\bf W}^{\intercal}{\bf x} + b$, where {\bf W} and $b$
are parameters we learn by fine-tuning the model on a dataset $D$.

We describe different designs for $\mathcal{A}$ as follows.

\noindent {\bf $\mathcal{A}_0$: Text-Pair Embedding}

We build a language model representation for pairs of members of the tuple, $x=\left(q,r,t\right)$ by simply  inputing them to Transformer models $\mathcal{B}$ in the standard sentence pair fashion. 
We consider four different configurations of $\mathcal{A}_0$, one for each following pair $\left(q,r\right)$, $\left(q,t\right)$, $\left(r,t\right)$, and one for the triplet, $\left(q, r, t\right)$, modeled as the concatenation of the previous three.
The representation for each pair is produced by a different and independent BERT instance, i.e., $\mathcal{B}_p$. More formally, we have the following three models $\mathcal{A}_0\left( \mathcal{B}_p(p)\right)$, $\forall  p \in \mathcal{D}_0$,
%
%
where $\mathcal{D}_0 = \{(q,r), (q,t), (r,t)\}$. Additionally, we design a model over $(q, r, t)$ with $\mathcal{A}_0\left( \cup_{p \in \mathcal{D}_0} \hspace{.3em}\mathcal{B}_p(p) \right)$, where $\cup$ means concatenation of the representations.
We do not use the short answer, $s$, as its contribution is minimal when using powerful Transformer-based models.

\noindent {\bf $\mathcal{A}_1$: Improved Text-Triple Embedding}

The models of the previous section are limited to pair representations. We improve this by designing $\mathcal{B}$ models that can capture pattern dependencies across $q$, $r$ and $t$.
To achieve this, we concatenate pairs of the three pieces of text above. 
We indicate this string concatenation with the $\circ$ operator.
Specifically, we consider $\mathcal{D}_1 = \{(q,r \circ t), (r,q \circ t), (t,q \circ r)\}$ and propose the following $\mathcal{A}_1$. As before, we have the individual models, $\mathcal{A}_1\left( \mathcal{B}_p(p)\right)$, $\forall  p \in \mathcal{D}_1$ as well as the combined model, $\mathcal{A}_1\left( \cup_{p \in \mathcal{D}_1} \hspace{.3em}\mathcal{B}_p(p) \right)$, where
%
%
again, we use different instances of $\mathcal{B}$ and fine-tune them together accordingly. 


\noindent {\bf $\mathcal{A}_2$: Peer Attention for Pair of Transformer-based Models}
Our previous designs instantiate different $\mathcal{B}$ for each pair, learning the feature representations of the target pair and the relations between its members, during the fine-tuning process.
This individual optimization prevents to capture patterns across the representations of different pairs as there is no strong connection between the $\mathcal{B}$ instances. Indeed, the combination of feature representations \emph{only} happens in the last classification layer.

\begin{figure}[t]
\center
\includegraphics[width=0.8\linewidth]{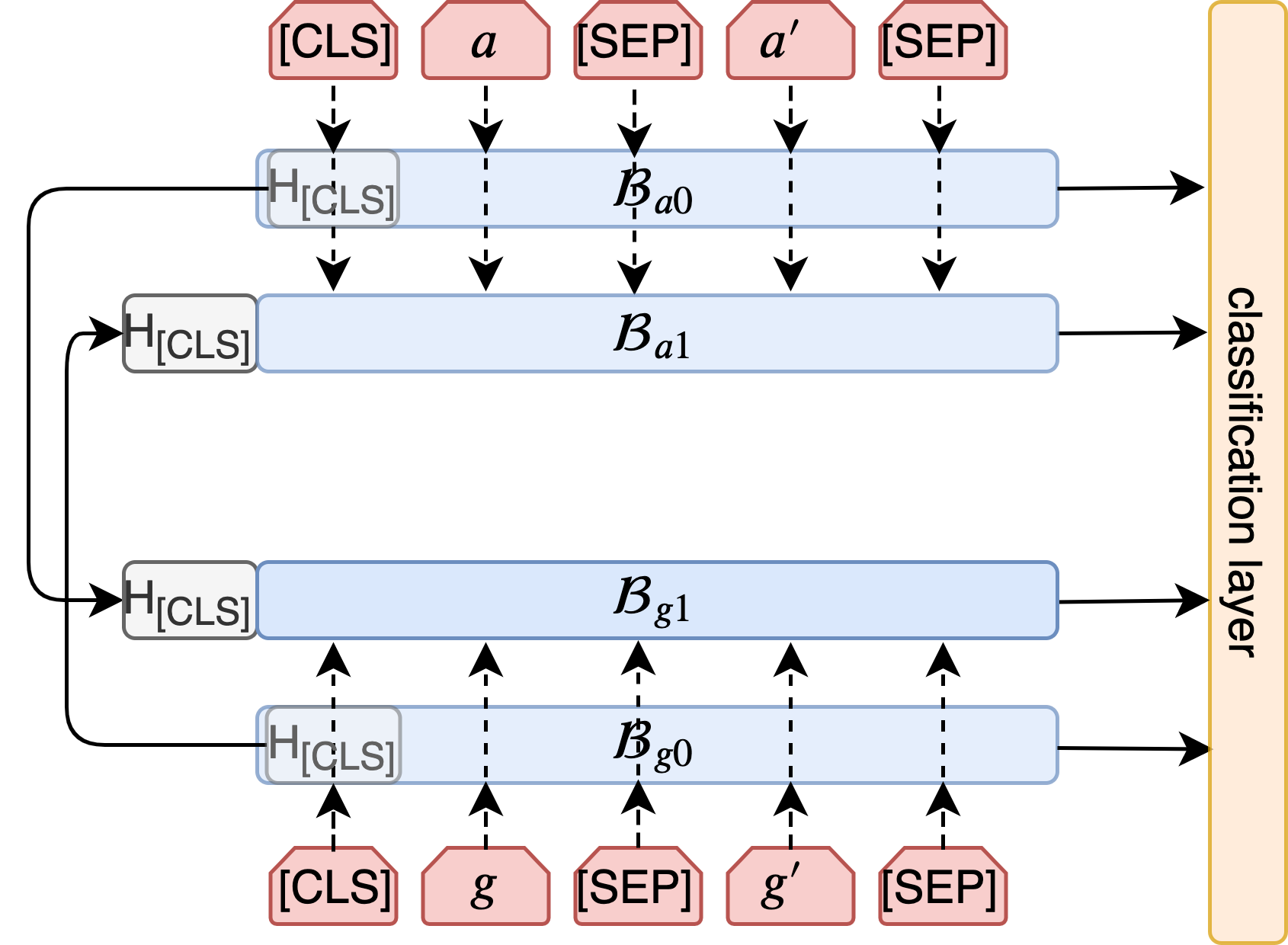}
\caption{peer attention on $\left(a, a'\right)$ and $\left(g, g'\right)$.}
\label{fig:peerattention}
\end{figure}

We propose \emph{peer-attention} to encourage the feature transferring between different $\mathcal{B}$ instances.
The idea, similar to encoder-decoder setting in Transformer-based models~\cite{NIPS2017_7181}, is to introduce an additional decoding step for each pair.
Figure~\ref{fig:peerattention} depicts our proposed setting for learning representation of two different pairs: $a_0=\left(a, a'\right)$ and $g_0=\left(g, g'\right)$.
The standard approach learns representations for these two in one pass, via $ \mathcal{B}_{a_0}$ and $\mathcal{B}_{g_0}$.
In \emph{peer-attention} setting, the representation output after processing one pair, captured in ${H}_{[CLS]}$, is input to the second pass of fine-tuning for the other pair.
Thus, the representation in one pair can attend over the representation in the other pair during the decoding stage. This allows the feature representations from each $\mathcal{B}$ instance to be shared both during  training and prediction stages.




\section{Dataset Creation}
\label{sec:data}

We describe the datasets we created to develop {\AVA}.
First, we build two large scale datasets for the standard QA task, namely {\bf {\qaNQ}} and {\bf {\qaGPD}}, derived from the Google Natural Questions dataset and our internal dataset, respectively. The construction of the datasets is described in Section~\ref{sec:qadata}.
Second, we describe our approach to generate labelled data for {\AVA} using the datasets for QA task, described in Section~\ref{sec:avadata}. 
Finally, we build an additional dataset constituted by a set of systems and their output on target test sets. This can be used to evaluate the ability of {\AVA} to estimate the end-to-end system performance (system-wise evaluation), described in Section~\ref{sec:evaldata}.

\subsection{Question Answering Datasets}
\label{sec:qadata}

\subsubsection{{\qaNQ}: {\ASS} Dataset from NQ}
\label{sec:qanq}

Google Natural Questions (NQ) is a large scale dataset for machine reading task~\cite{47761}.
Each question is associated with a Wikipedia page and at least one long paragraph (\texttt{long\_answer}) that contains the answer to the question.
The \texttt{long\_answer} may contain additional annotations of \texttt{short\_answer}, a succint extractive answer from the long paragraph. 
A \texttt{long\_answer} usually consists of multiple sentences, thus NQ is not directly applicable to our setting.

We create {\qaNQ} from NQ by leveraging
both \texttt{long\_answer} and \texttt{short\_answer} annotations. In particular for a given question, the (correct) answers for a question are sentences in the long answer paragraphs that contain \emph{annotated} \texttt{short\_answer}s. The other sentences from the Wikipedia page are considered incorrect.
The negative examples can be of the following types:
(i) Sentences that are in the \texttt{long\_answer} but do not contain \emph{annotated} short answers. It is possible that these sentences might contain the \texttt{short\_answer}.
(ii) Sentences that are not part of the \texttt{long\_answer} but contain a \texttt{short\_answer} as subphrase. Such occurrence is generally accidental.
(iii) All the other sentences in the document.

The generation of negative examples impacts on the robustness of the training model when selecting the correct answer out of the incorrect ones.  
{\qaNQ} has four labels that describe possible confusing levels of a sentence candidate.
We apply the same processing both to training and development sets of NQ.
This dataset enables to perform an effective transfer step~\cite{garg2019tanda}. Table~\ref{table:nq} shows the statistics of the dataset.

\begin{table*}
\centering
\resizebox{0.9\linewidth}{!}{%
\centering
\begin{tabular}{ c r r r|r r r|r r r } 
\toprule
									& \multicolumn{3}{c|}{\bf {\qaNQ}}                                                               & \multicolumn{3}{c|}{\bf {\qaNQ} Qs with multiple As}                                                                           & \multicolumn{3}{c }{\bf {\avaNQ}}                                                                    \\ 
\multicolumn{1}{ r }{data split}		& \multicolumn{1}{r }{\#Qs} & \multicolumn{1}{r }{\#As} & \multicolumn{1}{r|}{\#wrong-As} & \multicolumn{1}{r }{\#Qs} & \multicolumn{1}{r }{\#As} & \multicolumn{1}{l|}{\#wrong-As} & \multicolumn{1}{r }{positives} & \multicolumn{1}{r }{negatives} & \multicolumn{1}{r }{total}  \\ 
\hline
\multicolumn{1}{ r }{NQ-dev}   & 4,263                     & 134,691                   & 1,320,812                       & 1,478                                             & 3,376                     & 64,187                          & 11,556                         & 206,497                        & 218,053                     \\ 
\multicolumn{1}{ r }{NQ-train} & 105,020                   & 10,288                    & 33,294,803                      & 2,360                                             & 6,392                     & 96,152                          & 26,100                         & 432,913                        & 459,013                     \\
\bottomrule
\end{tabular}
}
\caption{{\qaNQ} and {\avaNQ} Statistics}
\label{table:nq}
\end{table*}

\subsubsection{{\qaGPD}: General Purpose Dataset}
\label{sec:qagpd}

A search engine using a large index can retrieve more relevant documents than those available in Wikipedia.
Thus, we retrieved high-probably relevant candidates as follows: we (i) retrieved top 500 relevant documents; (ii) automatically extracted the top 100 sentences ranked by a BERT model over all sentences of the documents; and (iii) had all the top 100 sentences manually annotated as correct or incorrect answers. 
This process does not guarantee that we have all correct answers but the probability to miss them is much lower than for other datasets. In addition, this dataset is richer than {\qaNQ} as it consists of answers from multiple sources. Furthermore, the average number of answers to a question is also higher than in {\qaNQ}.
Table~\ref{table:gpd} shows the statistics of the dataset.

\begin{table*}
\centering
\resizebox{0.9\linewidth}{!}{%
\centering
\begin{tabular}{ r r r r|r r r|r r r } 
\toprule
			& \multicolumn{3}{c|}{\bf {\qaGPD}}                                                              & \multicolumn{3}{c|}{\bf {\qaGPD} Qs with multiple As}                                                                           & \multicolumn{3}{c }{\bf {\avaGPD}}                                                                   \\ 
data split       & \multicolumn{1}{r }{\#Qs} & \multicolumn{1}{r }{\#As} & \multicolumn{1}{r|}{\#wrong-As} & \multicolumn{1}{r }{\#Qs} & \multicolumn{1}{r }{\#As} & \multicolumn{1}{r|}{\#wrong-As} & \multicolumn{1}{r }{positives} & \multicolumn{1}{r }{negatives} & \multicolumn{1}{r }{total}  \\ 
\hline
GPD-train & 262                       & 5,399                     & 20,801                          & 245                                               & 5,382                     & 20,748                          & 183,894                        & 349,765                        & 533,659                     \\ 
GPD-dev & 283                       & 8,682                     & 19,618                          & 276                                               & 8,674                     & 19,502                          & 430,230                        & 426,246                        & 856,476                     \\ 
GPD-test  & 294                       & 9,412                     & 19,988                          & 281                                               & 9,399                     & 19,790                          & 479,028                        & 449,625                        & 928,653                     \\
\bottomrule
\end{tabular}
}
\caption{{\qaGPD} and {\avaGPD} Statistics}
\label{table:gpd}
\end{table*}

\subsection{{\AVA} Datasets}
\label{sec:avadata}


The {\ASS} datasets from the previous section typically consist of a set of questions $Q$.
Each $q \in Q$ has $T_q=\{t_1, \dots, t_n\}$ candidates, comprised of both correct answers $C_q$ and incorrect answers $\overline{C_q}$, $T_q = C_q \cup \overline{C_q}$.
We construct the dataset for point-wise automatic evaluation (described in Section~\ref{sec:model}) in the following steps:
(i) to have positive and negative examples for {\AVA}, we first filter the QA dataset to only keep questions that have at least two correct answers. This is critical to build positive and negative examples.

Formally, let $\left\langle q, r, t, l \right\rangle$ be an input for {\AVA},
$$\text{AVA-Positives} = \left\langle q; \left( r, t \right) \in C_q \times C_q \text{ and } r \ne t \right\rangle$$
\noindent We also build negative examples as follows:
$$\text{AVA-Negatives} = \left\langle q; \left( r, t \right) \in C_q \times \overline{C_q} \right\rangle$$
We create {\avaNQ} and {\avaGPD} from the QA datasets, {\qaNQ} and {\qaGPD}.
The statistics are presented on the right side of tables~\ref{table:nq} and \ref{table:gpd}.

\subsection{{\AVA} Datasets from Systems (\avaADS)}
\label{sec:evaldata}

To test {\AVA} at level of overall system Accuracy, we need to have a sample of systems and their output on different test sets.
We create a dataset that has candidate answers collected from eight systems from a set of 1,340 questions. 
The questions were sampled from an anonymized set of user utterances.
We only considered information inquiry questions. The systems differ from each other in multiple ways, including: (i)~\emph{modeling}: Compare-Aggregate (CNN-based) and different Transformers-based architectures with different hyper-parameter settings; (ii)~\emph{training}: the systems trained on different resources; and (iii)~\emph{candidates}: the pool of candidates for the selected answers are different.

\section{Experiments}
\label{sec:experiments}

We study the following performance aspects of {\AVA} in predicting: (i) the correctness of the individual answers provided by systems to questions (point-wise estimation); and (ii) the overall system Accuracy. We evaluated QA Accuracy as well as passage reranking performance, in comparison with the  human labeling.

The first aspect studies the capacity of our different machine learning models, whereas the second provides a perspective on the practical use of {\AVA} to develop QA systems.

\begin{table}
\resizebox{\linewidth}{!}{%
\centering
\begin{tabular}{cl} 
\toprule
{\bf Model Setting}   & {\bf Configurations}                                          \\ 
\midrule
Linear Classifier & using 4 features $x_i$                                 \\ 
$\mathcal{A}_0$   & one for each and one for all from $\mathcal{D}_0$     \\ 
$\mathcal{A}_1$   & all possible combinations from $\mathcal{D}_1$         \\ 
$\mathcal{A}_2$   & the most probable setting from $\mathcal{A}_1$        \\
\bottomrule
\end{tabular}
}
\caption{The {\AVA} configurations used in training}
\label{table:configurations}
\end{table}

\subsection{Datasets}
We trained and test models using {\avaNQ} and {\avaGPD} datasets, described in Section~\ref{sec:avadata}. We also evaluate the point-wise performance on the WikiQA and TREC-QA datasets.

\subsection{Models}
Table~\ref{table:configurations} summarizes the configurations we consider for training and testing.
For the linear classifier baseline, we built a vanilla SVM classifier using~\texttt{scikit-learn}.
We set the \texttt{probability} parameter to enable Platt scaling calibration on the score of SVM.

We developed our Transformer-based evaluators on top of the HuggingFace's Transformer library~\cite{Wolf2019HuggingFacesTS}.
We use RoBERTa-Base as the initial pre-trained model for each $\mathcal{B}$ instance~\cite{DBLP:journals/corr/abs-1907-11692}.
We use the default hyperparameter setting for typical GLUE trainings.
This includes (i) the use of the AdamW variant~\cite{DBLP:journals/corr/abs-1711-05101} as optimizer, (ii) the learning rate of ~$1e\text{-}06$ set for all fine-tuning exercises, and (iii) the maximum sequence length set to 128.
The number of iterations is set to 2.
We also use a development set to enable early stopping based on F1 measure after the first iteration.
We fix the same batch size setting in the experiments to avoid possible performance discrepancies caused by different batch size settings.

\subsection{Metrics}
We study the performance of {\AVA} in evaluating passage reranker systems, which differ not only in methods but also in domains and application settings.
We employ the following evaluation strategies to benchmark {\AVA}.

\paragraph{Point-wise Evaluation} We study the performance of {\AVA} on point-wise estimation using traditional Precision, Recall, and F1.
The metrics indicate the performance of {\AVA} in predicting if an answer candidate is correct or not.

\paragraph{System-wise evaluation}
We measured {\AVA} when used in a simple aggregator to compute the overall system performance over a test set. The metrics we consider are:
Precision-at-1 (P@1), Mean Average Precision (MAP), and Mean Reciprocal Rank (MRR), when computing the performance on TREC-QA and WikiQA, since such datasets contain ranks of answers.
In contrast, we only use P@1 on {\avaADS} dataset, as this only includes the selected answers for each system.

We use Kendall's Tau-b\footnote{We use \texttt{scipy.stats.kendalltau}} to measure the correlation between the ranking produced by {\AVA} and the one available in the GS: 
$\tau=\frac{c-d}{c+d},$
where $c$ and $d$ are the numbers of concordant and discordant pairs between two rankings.
	
We additionally analyze the gap of each performance given by {\AVA} and the one computed with the GS, using root mean square error:
$\text{RMSE}\left(a, h\right) = \sqrt{\frac{1}{n}\Sigma_{i=1}^{n}{\left({a_i -h_i}\right)^2}},$
	where $a$ and $h$ are the measures given by {\AVA} and from human annotation respectively.
%

\subsection{Results on Point-wise Evaluation}

We evaluate the performance of {\AVA} in predicting if an answer $t$ is correct for a question $q$, given a reference $r$.
Table~\ref{table:pointwise} shows the result: Column 1 reports the names of the systems described in Section~\ref{sec:model}, while columns 2 and 3 show the F1 measured on {\avaNQ} and {\avaGPD}, respectively.

We note that: (i) the F1 on {\avaGPD} is much higher than the one on {\avaNQ}, this is because the former dataset is much larger than latter;\\ 
(ii) $\mathcal{A}_0\left(\{\left(q, r\right)\}\right)$ cannot predict if an answer is correct as it does not use it in the representation, thus its Accuracy is lower than 7\%;\\
(iii) $\mathcal{A}_0\left(\{\left(r, t\right)\}\right)$ is already a reasonable model mainly based on paraphrasing between $r$ and $t$;\\
(iv) $\mathcal{A}_0\left(\{\left(q, t\right)\}\right)$ is also a good model as it is as much powerful as a QA system;\\
(v) the $\mathcal{A}_1$ models that takes the entire triplet $q$, $r$ and $t$ are the most accurare achieving an F1 of almost 74\%;\\
(vi) the use of combinations of triplets, e.g., $\mathcal{A}_1\left(\{\left(r, q \circ t\right), \left(t, q \circ r\right)\}\right)$, provides an even more accurate model; and finally,\\
(vii) the \emph{peer-attention} model, i.e., $\mathcal{A}_2\left(\left(r, q\circ t\right), \left(t, q \circ r\right)\right)$ reaches almost 75\%. \vspace{-1em}
\begin{table}[t]
\resizebox{.9\linewidth}{!}{%
\centering
\begin{tabular}{ r r r } 
\toprule
training set from                         & \multicolumn{1}{c }{{\avaNQ}} & \multicolumn{1}{c }{{\avaGPD}}  \\ 
development set from                       & \multicolumn{1}{c }{{\avaNQ}}   & \multicolumn{1}{c }{{\avaGPD}}    \\ 
\midrule
\multicolumn{1}{ c }{\bf Model}        & \multicolumn{2}{c }{\bf F1 on {\avaGPD}-Test}                                              \\ 
\hline
\hline
Linear Classifier                  & \multicolumn{1}{r }{0.0000}            & 0.3999                             \\ 
\hline	
$\mathcal{A}_0\left(\{\left(q, r\right)\}\right)$& \multicolumn{1}{r }{0.0004}            & 0.0695                             \\ 
$\mathcal{A}_0\left(\{\left(r, t\right)\}\right)$ & \multicolumn{1}{r }{0.3778}            & 0.6247                             \\ 
$\mathcal{A}_0\left(\{\left(q, t\right)\}\right)$ & \multicolumn{1}{r }{0.5801}            & 0.6713                             \\ 
$\mathcal{A}_0\left(\mathcal{D}_0\right)$ & \multicolumn{1}{r }{0.3962}            & {\bf 0.6807}                            \\ 
\hline
\hline
$\mathcal{A}_1\left(\{\left(q, r \circ t\right)\}\right)$                         
& 0.3788                           & 0.7014                             \\ 
$\mathcal{A}_1\left(\{\left(r, q \circ t\right)\}\right)$                         
& 0.4583                           & 0.7383                             \\ 
$\mathcal{A}_1\left(\{\left(t, q \circ r\right)\}\right)$)                         
& 0.4517                           & 0.7236                             \\ 
\hline
$\mathcal{A}_1\left(\{\left(q, r \circ t\right), \left(t, q \circ r\right)\}\right)$
& 0.3546                           & 0.7421                             \\ 
$\mathcal{A}_1\left(\{\left(r, q \circ t\right), \left(t, q \circ r\right)\}\right)$
& 0.4002                           & {\bf 0.7447}                             \\ 
$\mathcal{A}_1\left(\{\left(r, q \circ t\right), \left(q, r \circ t\right)\}\right)$
& 0.4873                           & 0.7435                             \\ 
\hline
$\mathcal{A}_1\left(\mathcal{D}_1\right)$ & 0.4121                           & 0.7303                             \\ 
\hline
\hline
$\mathcal{A}_2\left(\left(r, q\circ t\right), \left(t, q \circ r\right)\right)$
& \multicolumn{1}{r }{0.4187}            & {\color{blue}\bf 0.7472}                             \\
\hline
\end{tabular}
}
\caption{F1 on {\avaGPD}-Test}
\label{table:pointwise}
\end{table}
\begin{table}[t]
\centering
\resizebox{1.00\linewidth}{!}{%
\begin{tabular}{ c c r r l } 
\toprule
\multicolumn{2}{ r}{\multirow{2}{*}{\bf Metrics}} & \multicolumn{2}{c }{\bf Kendall}                           & \multicolumn{1}{c }{\multirow{2}{*}{\bf RMSE$~\pm~\sigma$}}  \\ 
\multicolumn{2}{ c }{}                         & \multicolumn{1}{c }{$\tau$} & \multicolumn{1}{c }{$p$} & \multicolumn{1}{c }{}                       \\ 
\midrule
\multirow{3}{*}{TREC-QA-Dev}  & P@1            & 1.000                            & 0.003                       & 0.000$~\pm~$0.000                               \\ 
								& MAP            & 1.000                            & 0.003                       & 0.040$~\pm~$0.019                               \\ 
								& MRR            & 0.866                            & 0.017                       & 0.015$~\pm~$0.011                               \\ 
\hline
\multirow{3}{*}{TREC-QA-Test} & P@1            & 1.000                            & 0.003                       & 0.034$~\pm~$0.018                               \\ 
								& MAP            & 0.867                            & 0.017                       & 0.041$~\pm~$0.029                               \\ 
								& MRR            & 1.000                            & 0.003                       & 0.020$~\pm~$0.012                               \\ 
\hline
\hline
\multirow{3}{*}{WikiQA-Dev}  & P@1            & 1.000                            & 0.009                       & 0.000$~\pm~$0.000                               \\ 
								& MAP            & 0.733                            & 0.056                       & 0.050$~\pm~$0.039                               \\ 
								& MRR            & 0.690                            & 0.056                       & 0.063$~\pm~$0.052                               \\ 
\hline
\multirow{3}{*}{WikiQA-Test} & P@1            & 0.889                            & 0.017                       & 0.079$~\pm~$0.030                               \\ 
								& MAP            & 0.733                            & 0.056                       & 0.081$~\pm~$0.040                               \\ 
								& MRR            & 0.867                            & 0.017                       & 0.095$~\pm~$0.035                               \\
\bottomrule
\end{tabular}
}
\caption{System-wise evaluation on TREC-QA and WikiQA using AVA model, $\mathcal{A}_2\left(\left(r, q\circ t\right), \left(t, q \circ r\right)\right)$.}
\label{table:modeleval}
\end{table}
\begin{table}
\centering
\resizebox{1.05\linewidth}{!}{%
\begin{tabular}{ c c l r r r r r r } 
\toprule
								&                        & \multicolumn{1}{c }{\bf Metrics} & \multicolumn{1}{c }{\bf M1} & \multicolumn{1}{c }{\bf M2} & \multicolumn{1}{c }{\bf M3} & \multicolumn{1}{c }{\bf M4} & \multicolumn{1}{c }{\bf M5} & \multicolumn{1}{c }{\bf M6}  \\ 
\midrule
\multirow{6}{*}{\rotcell[cc]{TREC-QA-Dev}}  & \multirow{3}{*}{{\rotcell[cc]{Gold}}} & P@1                          & 0.717                   & 0.870                   & 0.891                   & 0.935                   & 0.739                   & 0.826                    \\ 
								&                        & MAP                          & 0.691                   & 0.858                   & 0.913                   & 0.912                   & 0.769                   & 0.796                    \\ 
								&                        & MRR                          & 0.819                   & 0.923                   & 0.937                   & 0.967                   & 0.835                   & 0.890                    \\ 
\cline{2-9}
								& \multirow{3}{*}{{\rotcell[cc]{\AVA}}}   & P@1                          & 0.717                   & 0.870                   & 0.891                   & 0.935                   & 0.739                   & 0.826                    \\ 
								&                        & MAP                          & 0.688                   & 0.831                   & 0.864                   & 0.857                   & 0.717                   & 0.772                    \\ 
								&                        & MRR                          & 0.809                   & 0.920                   & 0.940                   & 0.967                   & 0.803                   & 0.876                    \\ 
\hline
\multirow{6}{*}{\rotcell[cc]{Trec-QA-Test}} & \multirow{3}{*}{{\rotcell[cc]{Gold}}} & P@1                          & 0.596                   & 0.885                   & 0.904                   & 0.962                   & 0.712                   & 0.788                    \\ 
								&                        & MAP                          & 0.661                   & 0.873                   & 0.894                   & 0.904                   & 0.771                   & 0.801                    \\ 
								&                        & MRR                          & 0.763                   & 0.933                   & 0.945                   & 0.976                   & 0.820                   & 0.869                    \\ 
\cline{2-9}
								& \multirow{3}{*}{{\rotcell[cc]{\AVA}}}   & P@1                          & 0.635                   & 0.904                   & 0.962                   & 0.981                   & 0.712                   & 0.827                    \\ 
								&                        & MAP                          & 0.639                   & 0.845                   & 0.896                   & 0.886                   & 0.680                   & 0.789                    \\ 
								&                        & MRR                          & 0.764                   & 0.936                   & 0.981                   & 0.990                   & 0.793                   & 0.880                    \\ 
\midrule
\midrule
\multirow{6}{*}{\rotcell[cc]{WikiQA-Dev}}  & \multirow{3}{*}{{\rotcell[cc]{Gold}}} & P@1                          & 0.545                   & 0.727                   & 0.455                   & 0.545                   & 0.636                   & 0.727                    \\ 
								&                        & MAP                          & 0.636                   & 0.744                   & 0.656                   & 0.621                   & 0.755                   & 0.781                    \\ 
								&                        & MRR                          & 0.720                   & 0.831                   & 0.695                   & 0.703                   & 0.803                   & 0.864                    \\ 
\cline{2-9}
								& \multirow{3}{*}{{\rotcell[cc]{\AVA}}}   & P@1                          & 0.545                   & 0.727                   & 0.455                   & 0.545                   & 0.636                   & 0.727                    \\ 
								&                        & MAP                          & 0.523                   & 0.751                   & 0.643                   & 0.617                   & 0.713                   & 0.774                    \\ 
								&                        & MRR                          & 0.568                   & 0.841                   & 0.682                   & 0.698                   & 0.788                   & 0.841                    \\ 
\hline
\multirow{6}{*}{\rotcell[cc]{WikiQA-Test}} & \multirow{3}{*}{{\rotcell[cc]{Gold}}} & P@1                          & 0.563                   & 0.844                   & 0.781                   & 0.688                   & 0.813                   & 0.781                    \\ 
								&                        & MAP                          & 0.634                   & 0.778                   & 0.753                   & 0.746                   & 0.834                   & 0.820                    \\ 
								&                        & MRR                          & 0.746                   & 0.917                   & 0.876                   & 0.833                   & 0.906                   & 0.883                    \\ 
\cline{2-9}
								& \multirow{3}{*}{{\rotcell[cc]{\AVA}}}   & P@1                          & 0.625                   & 0.781                   & 0.719                   & 0.656                   & 0.719                   & 0.656                    \\ 
								&                        & MAP                          & 0.660                   & 0.750                   & 0.687                   & 0.683                   & 0.705                   & 0.704                    \\ 
								&                        & MRR                          & 0.732                   & 0.820                   & 0.783                   & 0.741                   & 0.791                   & 0.762                    \\
\bottomrule
\end{tabular}
}
\caption{Details of system-wise Evaluation on TREC-QA and WikiQA using AVA model and GS, $\mathcal{A}_2\left(\left(r, q\circ t\right), \left(t, q \circ r\right)\right)$.}
\label{table:modelevaldetails}
\end{table}
\begin{table*}
\centering
\small
\resizebox{1\linewidth}{!}{%
\begin{tabular}{lcrrrrrrrrrrc}
\toprule
\multirow{2}{*}{\bf ADS Split}   & \multirow{2}{*}{\bf Evaluator} & \multicolumn{1}{c}{\multirow{2}{*}{\bf S1}} & \multicolumn{1}{c}{\multirow{2}{*}{\bf S2}} & \multicolumn{1}{c}{\multirow{2}{*}{\bf S3}} & \multicolumn{1}{c}{\multirow{2}{*}{\bf S4}} & \multicolumn{1}{c}{\multirow{2}{*}{\bf S5}} & \multicolumn{1}{c}{\multirow{2}{*}{\bf S6}} & \multicolumn{1}{c}{\multirow{2}{*}{\bf S7}} & \multicolumn{1}{c}{\multirow{2}{*}{\bf S8}} & \multicolumn{2}{c}{\bf Kendall}                 & \multirow{2}{*}{\bf RMSE$~\pm~\sigma$}     \\
								&                            & \multicolumn{1}{l}{}                    & \multicolumn{1}{l}{}                    & \multicolumn{1}{l}{}                    & \multicolumn{1}{l}{}                    & \multicolumn{1}{r}{}                    & \multicolumn{1}{r}{}                    & \multicolumn{1}{l}{}                    & \multicolumn{1}{l}{}                    & \multicolumn{1}{c}{$\tau$}              & \multicolumn{1}{c}{$p$}                   &                           \\
\midrule
\multirow{2}{*}{Dev (20\%)}  & AVA                        & 0.215                                   & 0.278                                   & 0.22                                    & 0.369                                   & 0.285                                   & 0.294                                   & 0.283                                   & 0.355                                   & \multirow{2}{*}{0.929} & \multirow{2}{*}{0.0004}  & \multirow{2}{*}{0.0198$~\pm~$0.012}  \\
								& ADS                        & 0.218                                   & 0.282                                   & 0.234                                   & 0.379                                   & 0.309                                   & 0.315                                   & 0.261                                   & 0.319                                   &                          &                          &                           \\
\midrule
\multirow{2}{*}{Test (80\%)} & AVA                        & 0.235                                   & 0.289                                   & 0.235                                   & 0.355                                   & 0.319                                   & 0.321                                   & 0.301                                   & 0.357                                   & \multirow{2}{*}{0.643} & \multirow{2}{*}{0.031} & \multirow{2}{*}{0.0350$~\pm~$0.019}  \\
								& ADS                        & 0.235                                   & 0.324                                   & 0.26                                    & 0.393                                   & 0.356                                   & 0.365                                   & 0.249                                   & 0.336                                   &                          &                          &                          \\
\bottomrule
\end{tabular}
}
\caption{Details of system-wise Evaluation on {\avaADS} benchmark dataset}
\label{table:systemevaldetails}
\end{table*}
\begin{table*}
\centering
\resizebox{1\linewidth}{!}{%
\begin{tabular}{|p{0.24\linewidth}|p{0.45\linewidth}|c|p{0.45\linewidth}|c|} 
\hline
\multicolumn{1}{|c|}{\bf Question $q$}                                        & \multicolumn{1}{c|}{\bf Candidate $t$}                                                                                                                                                                                                                & \multicolumn{1}{l|}{\bf TANDA} & \multicolumn{1}{c|}{\bf Reference $r$}                                                                                                                                                    & \multicolumn{1}{c|}{\bf $\mathcal{A}$}  \\ 
\hline
when were the nobel prize awards first given ?                 & among them is the winner of the first prize in 1901 , sully prudhomme .                                                                                                   & 0.0001                                  & leo tolstoy lost the first literature prize in 1901 to the forgettable rene f . a . sully prudhomme .                                                                                                                                 & 0.596                           \\ 
\hline
what branch of the service did eileen marie collins serve in ? & the first woman to command a space shuttle mission , air force col . eileen collins , sees her flight next month as \`{}\`{} a great challenge '' in more ways than one . & 0.046                                  & shuttle commander eileen collins , a working mother and air force colonel , was set to make history as the first woman to command a space mission .                                                                                   & 0.895                           \\ 
\hline
what was johnny appleseed 's real name ?                       & appleseed , whose real name was john chapman , planted many trees in the early 1800s .                                                                                    & 0.026                                  & whitmore said he was most fascinated with the story of john chapman , who is better known as johnny appleseed .                                                                                                                       & 0.948                           \\ 
\hline
when was the challenger space shuttle disaster ?               & sept . 29 , 1988 \_ americans return to space aboard the shuttle discovery , after a 32-month absence in the wake of the challenger accident .                       & 0.995                                  & challenger was lost on its 10th mission during a 1986 launch accident that killed seven crew members .                                                                                                                       & 0.080                           \\ 
\hline
when did jack welch become chairman of general electric ?      & everyone knew it was coming , but now they know when : john f . welch jr . , the chairman of general electric , will retire after the company 's annual meeting in april 2001 .                                                                                    & 0.968                                  & welch has turned what had been a \$ 25 billion manufacturing company in 1981 into a \$ 100 billion behemoth that derives huge portions of its revenues from more profitable services .                                                                                                                       & 0.064                           \\ 
\hline
	\end{tabular}
}
\caption{Examples show AVA can detect the failures of the State-of-the-art model by~\citet{garg2019tanda}.}
\label{table:qualitative}
\end{table*}
\subsection{Results on system-wise evaluation}
We evaluate the ability of {\AVA} in predicting the Accuracy of QA systems as well as the performance of answer sentence reranking tasks.
We conduct two evaluation studies with two public datasets, TREC-QA and WikiQA, and an internal {\avaADS} dataset.
\subsubsection{Results on public datasets}
For TREC-QA and WikiQA, we used a bag of different models against the development and test sets and compared the results with the performance measured by {\AVA} using one of the best model according to the point-wise evaluation, i.e., $\mathcal{A}_2\left(\left(r, q\circ t\right), \left(t, q \circ r\right)\right)$.

More specifically, we apply each model $m$ to select the best answer $t$ from the list of candidates for $q$ in the dataset. We first compute the performance of model $m$ based on the provided annotations.
The metrics include Accuracy or Precision-at-1 (P@1), MAP, and MRR.
We then run {\AVA} for $\left(q, t\right)$ using the GS answers of $q$ as reference $r$.
The final {\AVA} score is the average of {\AVA} scores applied to different references for $q$.
Before computing the Accuracy on the test set, we tune the {\AVA} threshold to minimize the RMSE between the Accuracy (P@1) measured by {\AVA} and the one computed with the GS, on the development set of each dataset. We use these thresholds to evaluate the results on the test sets.

We considered six different models, including one Compare-Aggregate (CNN) trained model and five other Transformers-based models. Four of the latter are collected from public resources\footnote{\texttt{github.com/alexa/wqa\_tanda}} \cite{garg2019tanda}.
These models differ in the architectures and their training data thus their output is rather different.
We removed questions that have no correct or no incorrect answers.

Table~\ref{table:modeleval} reports the overall results averaged over the six models. We note that (i) setting the right threshold on the dev.~set, the error on P@1 is 0; (ii) this is not the case for MAP, which is a much harder value to predict as it requires to estimate an entire ranking; (iii) on the TREC-QA test set, AVA has an error ranging from 2 to 4.1 points on any measure; (iv) on the WikiQA test set, the error is higher, reaching 10\%, probably due to a larger complexity of the questions; (v) the std.~dev. is low, suggesting that AVA can be used to estimate system performance.

Additionally, we compute the Kendall's Tau-b correlation between the ranking of the six systems sorted in order of performance (P@1) according to the GS and AVA. We observe a perfect correlation on TREC-QA and a rather high correlation on WikiQA. This means that AVA can be used to determine if a model is better than another, which is desirable when developing new systems. The low p-values indicate reliability of our results. 


Finally, Table~\ref{table:modelevaldetails} shows the comparison between the performance evaluated with GS (Human) and AVA for all six models. The predictions of AVA are close to those from human judgement.

\subsubsection{Results on {\avaADS}}
We use {\avaADS}  dataset in this evaluation.
The task is more challenging as {\AVA} only receives one best answer for a system selected from different candidate pools.
There was also no control of the sources for the candidates.
Table~\ref{table:systemevaldetails} shows the result. We note a lower correlation due to the fact that the 8 evaluated systems have very close Accuracy. On the other hand, the RMSE is rather low 3.1\% and the std.~dev. is also acceptable $<0.02$, suggesting an error less than 7\% with a probability $>$ 95\%.

\subsection{Qualitative Analysis}
Table~\ref{table:qualitative} reports some example questions from TREC-QA test set, the top candidate selected by the TANDA system~\cite{garg2019tanda}, the classification score of the latter, and the AVA score.
AVA judges an answer correct if the score is larger than 0.5.
We note that even if the score of TANDA system is low, AVA assigns to the answer a very high score, indicating that it is correct (see the first three examples).
Conversely, a wrong answer could be classified as such by AVA, even if TANDA assigned it a very large score (see the last two examples).

\section{Conclusion}

We presented {\AVA}, an automatic evaluator method for QA systems.
Specifically, we discussed our data collection strategy and model design to enable {\AVA} development.
First, we collected seven different datasets, classified into three different types, which we used to develop {\AVA} in different stages.
Second, we proposed different Transformer-based modeling designs of {\AVA} to exploit the feature signals relevant to address the problem.
Our extensive experimentation has shown the effectiveness of {\AVA} for different types of evaluation: point-wise and system-wise over Accuracy, MAP and MRR.


%

\bibliography{ava}
\bibliographystyle{acl_natbib}



\end{document}